\documentclass[letterpaper, 10 pt, conference]{ieeeconf}  

\IEEEoverridecommandlockouts                              

\usepackage{graphics} 
\usepackage{epsfig} 
\usepackage{amsmath} 
\usepackage{amssymb}  
\usepackage{cite}
\usepackage{bm}
\usepackage{hyperref}

\title{\LARGE \bf
ATRos: Learning Energy-Efficient Agile Locomotion for Wheeled-legged Robots
}

\author{Jingyuan Sun$^{1*}$, Hongyu Ji$^{2* \dagger}$, Zihan Qu$^{3 \dagger}$, Chaoran Wang$^{4 \dagger}$, Mingyu Zhang$^{5 \dagger}$ 
\thanks {$^{*}$These authors contributed equally to this work.}
\thanks{\(^{\dagger}\)This work was performed during an internship at Huawei.}
\thanks{$^{1}$Jingyuan Sun is with Shanghai Huawei Technologies Co., Ltd., Shanghai 201799, China (corresponding author to provide phone:
18603680666; fax: none; e-mail:{\tt\small sunjingyuan1@huawei.com}).}%
\thanks{$^{2}$Hongyu Ji is with the College of Future Information Technology, Fudan University, Shanghai 200433, China, {\tt\small hyji21@m.fudan.edu.cn}}%
\thanks{$^{3}$Zihan Qu is with the Robotics Institute, Carnegie Mellon University, Pittsburgh, PA 15213, USA.} 
\thanks{$^{4}$Chaoran Wang is with the School of Aeronautic and Astronautics, Zhejiang University, Hangzhou 310027, China.} 
\thanks{$^{5}$Mingyu Zhang is with the School of Electronics and Communication Engineering, Sun Yat-sen University, Shenzhen 518107, China.}%
}

\begin{document}

\maketitle
\thispagestyle{empty}
\pagestyle{empty}

\begin{abstract}

Hybrid locomotion of wheeled-legged robots has recently attracted increasing attention due to their advantages of combining the agility of legged locomotion and the efficiency of wheeled motion.
But along with expanded performance, the whole-body control of wheeled-legged robots remains challenging for hybrid locomotion.
In this paper, we present \textbf{\textit{ATRos}}, a reinforcement learning (RL)-based hybrid locomotion framework to achieve hybrid walking-driving motions on the wheeled-legged robot. Without giving predefined gait patterns, our planner aims to intelligently coordinate simultaneous wheel and leg movements, thereby achieving improved terrain adaptability and improved energy efficiency. 
Based on RL techniques, our approach constructs a prediction policy network that could estimate external environmental states from proprioceptive sensory information, and the outputs are then fed into an actor-critic network to produce optimal joint commands. 
The feasibility of the proposed framework is validated through both simulations and real-world experiments across diverse terrains, including flat ground, stairs, and grassy surfaces.
The hybrid locomotion framework shows robust performance over various unseen terrains, highlighting its generalization capability. The details about real-world experiments can check (\url{https://baoziweiyuebing.github.io/ATRos-Wheeled_legged-robot-hybrid-locomotion/})

\end{abstract}

\section{INTRODUCTION}

Robots capable of agile and robust locomotion are becoming increasingly important in real-world scenarios such as industrial inspection, warehouse automation, urban delivery, and disaster response. 
While legged robots have demonstrated remarkable progress in traversing unstructured terrains, they often suffer from relatively high energy consumption and limited efficiency during long-range navigation. 
Wheeled robots, on the other hand, offer high-speed and energy-efficient mobility on structured surfaces but lack the versatility to overcome obstacles and operate in irregular environments. 
Wheeled-legged robots have recently emerged as a promising solution, combining the best of both paradigms by enabling rolling on flat terrain and stepping when required to overcome obstacles. 
This hybrid locomotion capability not only increases operational versatility but also introduces new opportunities for energy-efficient, long-duration missions.


Despite recent progress, hybrid locomotion for wheeled–legged robots remains highly challenging. Compared with purely legged systems, wheeled–legged platforms must cope with additional dynamics introduced by continuous wheel rolling, nonholonomic constraints, and frequent transitions between rolling and stepping modes. These robots also present a distinct energy–agility trade-off: wheeled locomotion on flat or moderately uneven terrain is typically more energy-efficient, as continuous wheel rotation demands less power than repetitive leg lifting. In contrast, negotiating obstacles or climbing stairs requires legged stepping, which is substantially more energy-intensive, particularly for heavier platforms. 
This characteristic further complicates the design of hybrid locomotion frameworks, as it necessitates the coordination of simultaneous walking and driving behaviors while ensuring agility, versatility, and energy efficiency.

Recent advances in reinforcement learning (RL) have enabled robots to learn sophisticated locomotion strategies directly from trial-and-error interactions, bypassing the need for handcrafted gait design. 
RL has shown strong performance on quadrupedal robots, generating agile, adaptive, and robust locomotion behaviors. 
However, directly transferring such approaches to wheeled-legged robots is non-trivial due to the expanded action space, modified contact dynamics, and the need for explicit energy-awareness during training. 
Without careful adaptation, learned policies tend to favor either overly aggressive locomotion, which wastes energy, or overly conservative behaviors, which sacrifice agility.

To address these challenges, we present \textbf{\textit{ATRos}}, 
a reinforcement learning framework for agile and transferable locomotion on wheeled-legged robots with a focus on energy efficiency. 
ATRos extends RL-based locomotion learning by incorporating rolling dynamics into the control architecture and introducing an energy-aware reward design that explicitly penalizes excessive power consumption while rewarding agile, stable motion. 
This enables the robot to discover hybrid locomotion strategies that achieve both efficiency and versatility, leveraging wheels for sustained rolling and legs for robust stepping when necessary. 

In summary, the contributions of this paper are:
\begin{itemize}
    \item We highlight the unique challenges of training locomotion policies for wheeled-legged robots compared to legged systems, focusing on the trade-off between agility and energy efficiency.
    \item We introduce ATRos, a reinforcement learning framework that achieves locomotion control on wheeled-legged platforms using only proprioceptive information, with an energy-aware reward design.
    \item We validate ATRos in both simulation and preliminary real-world experiments on a wheeled-legged quadruped robot, demonstrating improved energy efficiency and agile locomotion.
\end{itemize}

\section{Related Work}

\subsection{Hybrid Locomotion of Wheeled-Legged Robots}
Wheeled locomotion has been extensively studied in both laboratory and commercial applications, benefiting from its simplicity in control and high energy efficiency.
In contrast, legged locomotion remains more challenging due to the high degrees of freedom and the indirect nature of base motion through intermittent foot-ground contacts, which has motivated heuristic, bio-inspired, and optimization-based strategies. Wheeled-legged robots aim to combine the efficiency of wheels with the versatility of legs. However, research on fully integrated walking-driving motion remains relatively limited. Early efforts such as DRC-HUBO+~\cite{lim2017robot} and RoboSimian~\cite{hebert2015mobile} primarily exploited wheels, with legs playing only auxiliary roles. 
More advanced system likes 
ANYmal with wheels~\cite{Bjelonic2019} demonstrated driving-stepping capabilities for stairs and obstacles, highlighting the potential of hybrid designs. 
More recently, the FLORES project~\cite{song2025flores} proposed a redesigned wheel-leg architecture coupled with a tailored RL controller, enabling smooth mode transitions and improved adaptability across diverse terrains. Despite these advances, designing controllers that can seamlessly coordinate rolling and stepping while balancing agility, versatility, and energy efficiency remains an open challenge.

\subsection{Learning Approaches for Locomotion}

Although optimization-based methods have made notable progress in locomotion control, their reliance on simplified dynamics models often results in a trade-off between accuracy and computational efficiency~\cite{choi2023learning}. 
These limitations have motivated a growing interest in learning-based approaches, which allow robots to acquire complex behaviors through a direct function approximation. 
Reinforcement learning algorithms such as PPO~\cite{schulman2017proximal} have been widely adopted for 3D locomotion,
and hierarchical DRL frameworks further extend this by decomposing high-level motion planning and low-level control, enabling long-horizon tasks and multi-behavior training~\cite{tsounis2019deepgait}. 
Recent advances also emphasize robustness and sim-to-real transfer. Lee \textit{et al.}~\cite{lee2020learning} developed a blind locomotion controller for rough terrain, while Hwangbo \textit{et al.}~\cite{Hwangbo2019} demonstrated actuator network training that effectively transfers to real quadrupeds. 
Building on reinforcement learning, \cite{lee2024robust} introduced a hierarchical RL framework that integrates locomotion control with long-horizon navigation planning, successfully deployed in urban missions of the kilometer scale.
Despite these successes, learning approaches still face challenges in achieving reliable real-world deployment across diverse environments.

\section{Wheeled-Legged Locomotion Policy}
\subsection{Framework Overview}

        \begin{figure}[t]
		\centering
		\includegraphics
        [width=85mm]{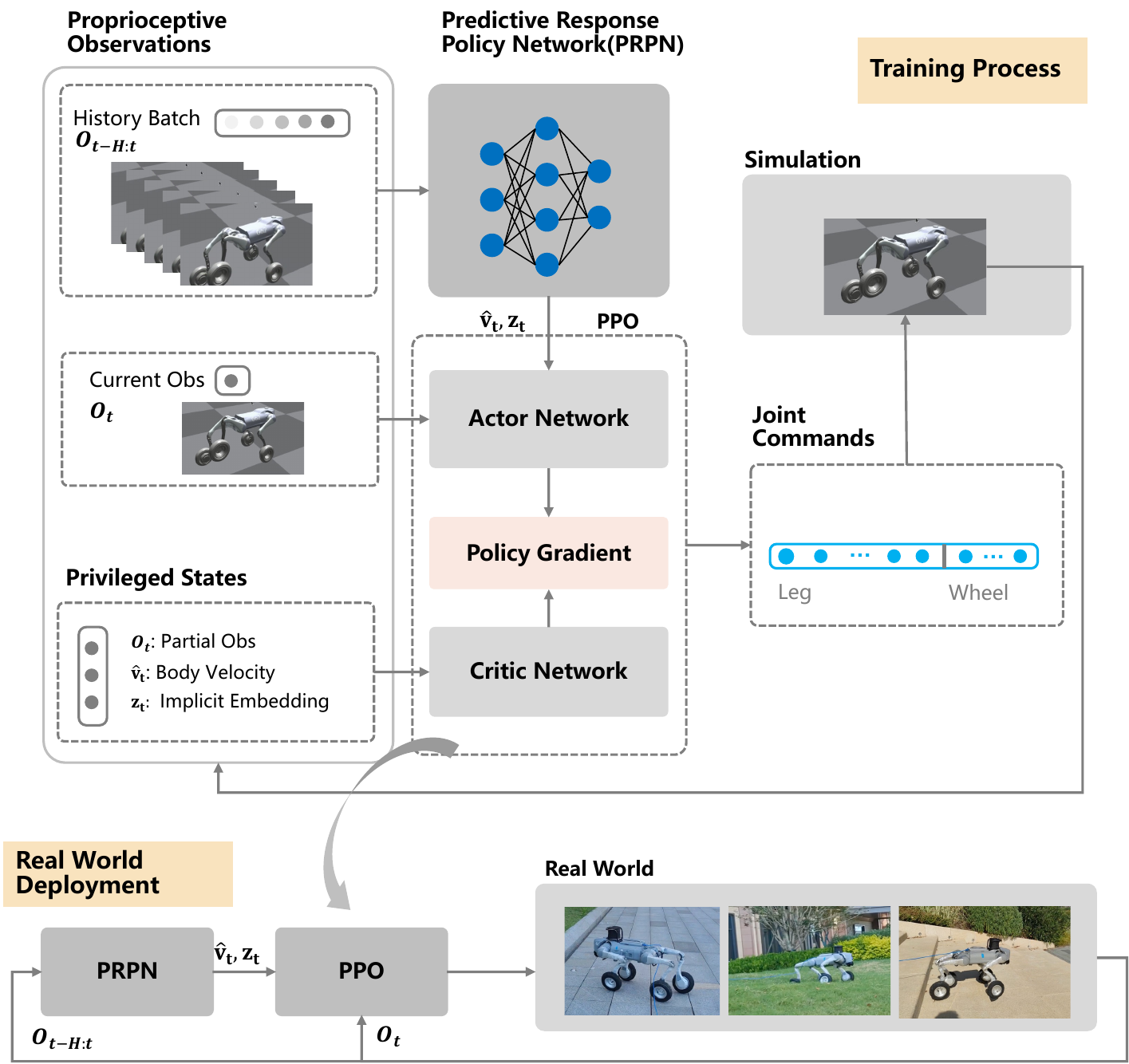}
		\caption{The overview framework of \textbf{\textit{ATRos}}. 
        PRPN encodes partial observations into predictive signals for a PPO-based actor–critic framework, enabling adaptive and energy-efficient locomotion with zero-shot sim-to-real transfer.
        }
		\label{fig:overview}
	\end{figure}

    
The wheeled-legged locomotion problem is formulated as a partially observable Markov decision process (POMDP),
since critical factors such as terrain variations and external forces are not directly observable. At each timestep $t$, the agent receives a partial observation $\mathbf{o}_t \in \mathcal{O}$ and generates an action $\mathbf{a}_t \sim \bm{\pi}_\theta(\mathbf{a}_t | \mathbf{o}_t, \hat{\mathbf{v}}_t, \mathbf{z}_t)$ using the policy $\bm{\pi}_\theta$. To mitigate partial observability, we introduce a \textit{Predictive Response Policy Network} (PRPN), which encodes a history of observations $\mathbf{o}_{t-H:t}$ to predict the body velocity $\hat{\mathbf{v}}_t$ and an implicit embedding $\mathbf{z}_t$, where $H$ represents the number of the history batch steps. These predictive signals, together with the current observation $\mathbf{o}_t$, are passed to the actor network, which outputs the action $\mathbf{a}_t$. The critic network, in turn, receives both the privileged state $\mathbf{s}_t$
and the current observation $\mathbf{o}_t$ to provide value estimates for policy updates. 

The actor and critic are jointly trained with proximal policy optimization (PPO)~\cite{schulman2017proximal} to maximize the expected return while ensuring stable optimization. 
Our formulation explicitly integrates predictive dynamics $\hat{\mathbf{v}}_t$ and latent embeddings $\mathbf{z}_t$ as auxiliary signals. This design provides three advantages: (i) predictive modeling of body dynamics improves locomotion stability, (ii) implicit embeddings $\mathbf{z}_t$ offer a compact and transferable state representation for robust policy learning, and (iii) leveraging privileged information in the critic accelerates training and improves sample efficiency for sim-to-real transfer.

        \begin{figure*}[t]
		\centering
		\includegraphics
        [width=1\textwidth]
        {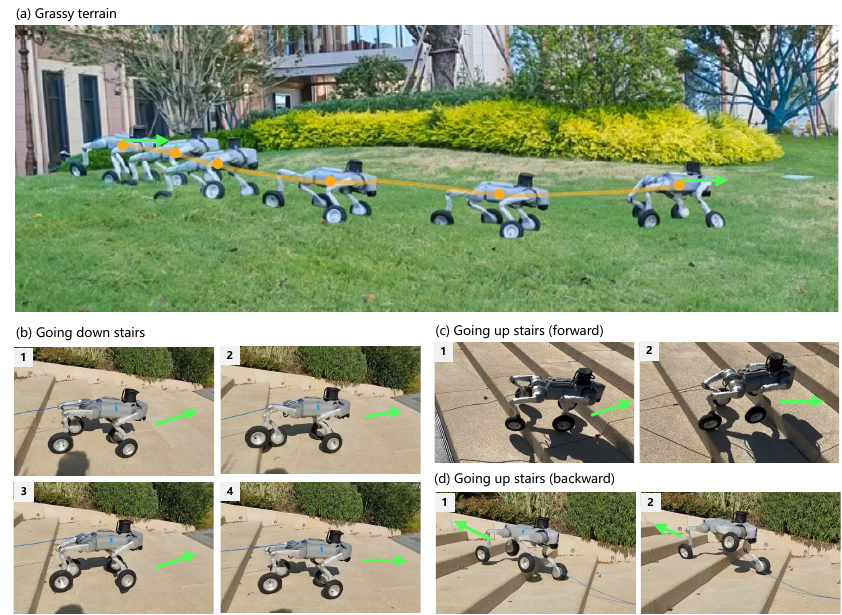} 
		\caption{
        Snapshots of Go2W robot crossing different types of terrains: grass and stairs with the height of 13 cm. (a) Wheel-based traversal of grassy terrain and small slopes. (b) Wheel-based traversal of stair descent. (c) Forward stair ascent achieved through wheeled-legged locomotion. (d) Backward stair ascent through wheeled-legged locomotion.
        }
		\label{fig:real_exp}
	\end{figure*}

\textbf{State Space.} 
The state $s_t$ is composed of proprioceptive observations:

\begin{equation}
\mathbf{s}_{t} = \left [\mathbf{c}_{t}, ~\mathbf{ \omega}_{t}, ~\mathbf{g}_{t}, ~\mathbf{q}_{t}, ~\mathbf{\dot{q}}_{t}, ~\mathbf{a}_{t-1} \right ]
\label{equa:proprioceptive observations}
\end{equation}
where $\mathbf{c}_{t} \in \mathbb{R}^{3}$ represents the body velocity command, $~\mathbf{ \omega}_{t} \in \mathbb{R}^{3}$ indicates the body angular velocity, $\mathbf{g}_{t} \in \mathbb{R}^{3}$ is the gravity vector, $\mathbf{q}_{t} \in \mathbb{R}^{16}$ and $\mathbf{\dot{q}}_{t} \in \mathbb{R}^{16}$ are the joint positions and joint velocities, $\mathbf{a}_{t-1} \in \mathbb{R}^{16}$ shows the previous action.
Thus, the proprioceptive state dimension is $\mathbf{s}_{t} \in \mathbb{R}^{57}$.


%
\textbf{Action Space.}
Different from pure legged locomotion, the policy here outputs actions for both legs $\mathbf{a}_{t}^{\mathrm{leg}} \in \mathbb{R}^{12}$ and wheels $\mathbf{a}_{t}^{\mathrm{wheel}} \in \mathbb{R}^{4}$.
To accelerate the training process,
the legged action space $\mathbf{a}_{t}^{\mathrm{leg}}$ is scaled by a factor $k_{l}$ to constrain the output action and improve stability, and it is represented as an increment relative to a reference joint configuration.
Thus, the target joint position of legs can be computed as:
\begin{equation}
q^{\text{des}}_t = k_{l} \mathbf{a}_{t}^{\mathrm{leg}} + q^{\text{ref}}_t,
\end{equation}
where $q^{\text{ref}}_t$ is the reference joint configuration in the robot stand pose. 
In the wheeled action space, $k_{w} \mathbf{a}_{t}^{\mathrm{wheel}}$ is interpreted as the target velocity.

\textbf{Reward Function.}
The overall reward is formulated as a weighted combination of multiple sub-rewards:
\begin{equation}
\mathbf{r}_{t} \left( \mathbf{s}_{t}, \mathbf{a}_{t} \right ) = \sum{\alpha_{i}r_{i}},
\end{equation}
where $\alpha_{i}$ denotes the weight coefficient assigned to each reward component, and $r_{i}$ represents a specific reward term. We adopt the reward structure in \cite{HIM} and additionally design wheel-related reward functions to better capture the characteristics of wheeled-legged locomotion. This extended reward formulation promotes efficient, stable, and terrain-adaptive locomotion, thereby facilitating robust sim-to-real transfer. 

\subsection{Training Details}
To effectively train robust locomotion policies for wheeled-legged robots, we adopt a  simulation framework combined with domain randomization and curriculum learning. The details are as follows.

\begin{itemize}
    \item Simulation Setup: We train policies in Isaac Gym~\cite{rudin2022learning} using 4096 parallel environments with a rollout horizon of 100 steps. The framework allows fast, large-scale data collection, and stable performance is typically reached after several thousand iterations.
    \item Dynamics Randomization: To improve robustness and facilitate sim-to-real transfer, we randomize key physical parameters of both the robot and environment, including body and link masses, center of mass, payloads, ground properties, motor strength, PD gains, actuation delay, external perturbations, and initial configurations.
    \item Training Curriculum. Following~\cite{rudin2022learning}, we employ a terrain curriculum to train the hybrid locomotion policy. Robots are initially trained on simple terrains, with difficulty gradually increasing based on performance. The terrain set contains 200 height field-based variations arranged by difficulty, and the command space is progressively expanded to cover complex locomotion scenarios such as slopes, stairs, and uneven grounds.
\end{itemize}

\section{Experiments}
In this section, we evaluate the proposed \textbf{\textit{ATRos}} framework both in simulation and real-world environments.  
The experiments are designed to assess locomotion agility, stability, and energy efficiency across two representative terrains: (i) grass terrain, (ii) stairs. 
The details of the experimental setup and details are introduced as following.

\subsection{Experimental setup}
For hardware deployment, we use a Unitree Go2W EDU wheeled-legged robot, which possesses 16 DoFs (twelve for legged actuators and four for wheeled actuators) as shown in Fig.~\ref{fig:real_exp}. The computation is provided by an NVIDIA RTX 4090 GPU, which enables real-time operation. The RL-based hybrid locomotion policy runs at 50 Hz, ensuring responsive execution of leg and wheel actions. The generated joint torques are transmitted from the computer to the robot actuators via a wired Ethernet connection, supporting low-latency and reliable deployment in physical environments.



\subsection{Cross grass terrains}
\label{sec:exp-grass terrain}
The grass terrain consist of soft, uneven outdoor grass-
land with natural slopes and scattered pits, designed to
test adaptability, stability, and robustness under irregular
conditions. 
On grass terrain, the robot tends to utilize its wheels for rolling, which highlights the energy efficiency of wheeled locomotion while simultaneously maintaining motion agility, thus achieving a balance between efficiency and flexibility. 
The snapshots of these experiments are shown in Fig.~\ref{fig:real_exp}(a).



\subsection{Cross stairs}
\label{sec:exp-cross stairs}

To evaluate the obstacle adaptability performance of the hybrid locomotion system, ascending and descending experiments were carried out on an outdoor staircase with a step height of 13 cm. It was anticipated that during descent, the robot would predominantly use wheeled motion to improve energy efficiency, whereas during ascent, the legged versatility would be fully utilized to enhance obstacle traversal.

As illustrated in the accompanying figure, the robot employed exclusively wheeled locomotion throughout the entire descending phase, as shown in the test snapshot in Fig.~\ref{fig:real_exp}(b). During ascent, 
the robot effectively leveraged the combined wheeled-legged locomotion, 
with the corresponding test snapshot provided in Fig.~\ref{fig:real_exp}(c).

Furthermore, the robot is tested in a reverse stair-ascent mode. The results demonstrate that the robot maintains excellent performance even when ascending stairs backward, successfully achieving stable motion. A snapshot from this experiment is presented in Fig.~\ref{fig:real_exp}(d).



\section{CONCLUSION}
In this paper, a hybrid locomotion framework is proposed to address the coordinated collaboration between wheels and legs for realizing smooth locomotion. The experiments demonstrate that ATRos achieves both agile and energy-efficient locomotion over diverse terrains. ATRos prefers wheeled mode that significantly realizing energy efficient motions on rolling-dominated terrains while maintaining comparable agility on uneven and stair environments.

Our future work will focus on two main aspects. First, we aim to enhance the stability of obstacle negotiation. Then, we are going to improve the integration of wheels and legs to achieve more natural and reliable wheeled-legged motions.

\addtolength{\textheight}{-12cm}   





\bibliographystyle{IEEEtran}
\bibliography{bib/sn-bibliography}

\end{document}